\documentclass[letterpaper]{article} 
\usepackage{aaai23}  
\usepackage{times}  
\usepackage{helvet}  
\usepackage{courier}  
\usepackage[hyphens]{url}  
\usepackage{graphicx} 
\usepackage{natbib}  
\usepackage{caption} 
\usepackage{algorithm}
\usepackage{algorithmic}

\usepackage{amssymb}
\usepackage{amsmath}

\usepackage{booktabs}
\usepackage{multirow}

\usepackage{newfloat}
\usepackage{listings}
\DeclareCaptionStyle{ruled}{labelfont=normalfont,labelsep=colon,strut=off} 
\floatstyle{ruled}
\newfloat{listing}{tb}{lst}{}
\floatname{listing}{Listing}
\title{Towards Real-Time Panoptic Narrative Grounding by an End-to-End \\ Grounding Network}
\author{
    Haowei Wang\textsuperscript{\rm 1}\equalcontrib,
    Jiayi Ji\textsuperscript{\rm 1}\equalcontrib, 
    Yiyi Zhou\textsuperscript{\rm 1, 2}, 
    Yongjian Wu\textsuperscript{\rm 4}, 
    Xiaoshuai Sun\textsuperscript{\rm 1, 2, 3}\thanks{The corresponding author.}
}
\affiliations{
    \textsuperscript{\rm 1}Media Analytics and Computing Lab, Department of Artificial Intelligence, \\
    School of Informatics, Xiamen University, 361005, China. 
    
    \textsuperscript{\rm 2}Institute of Artificial Intelligence, Xiamen University, China.     
    
    \textsuperscript{\rm 3}Fujian Engineering Research Center of Trusted Artificial Intelligence Analysis and Application, Xiamen University, China.     
    
    \textsuperscript{\rm 4}Tencent Youtu Lab, Shanghai, China

    wanghaowei@stu.xmu.edu.cn, jjyxmu@gmail.com, zhouyiyi@xmu.edu.cn, \\
    littlekenwu@tencent.com, xssun@xmu.edu.cn
}

\begin{document}

\maketitle

\begin{abstract}
Panoptic Narrative Grounding (PNG) is an emerging cross-modal grounding task, which locates the target regions of an image corresponding to the text description.
Existing approaches for PNG are mainly based on a two-stage paradigm, which is computationally expensive. In this paper, we propose a one-stage network for real-time PNG, termed End-to-End Panoptic Narrative Grounding network (EPNG), which directly generates masks for referents. Specifically, we propose two innovative designs, \emph{i.e.}, Locality-Perceptive Attention (LPA) and a bidirectional Semantic Alignment Loss (SAL), to properly handle the many-to-many relationship between textual expressions and visual objects. LPA embeds the local spatial priors into attention modeling, \emph{i.e.}, a pixel may belong to multiple masks at different scales, thereby improving segmentation. To help understand the complex semantic relationships, SAL proposes a bidirectional contrastive objective to regularize the semantic consistency inter modalities. Extensive experiments on the PNG benchmark dataset demonstrate the effectiveness and efficiency of our method. Compared to the single-stage baseline, our method achieves a significant improvement of up to 9.4\% accuracy. More importantly, our EPNG is 10 times faster than the two-stage model. Meanwhile, the generalization ability of EPNG is also validated by zero-shot experiments on other grounding tasks. The source codes and trained models for all our experiments are publicly available at \url{https://github.com/Mr-Neko/EPNG.git}.
\end{abstract}

\section{Introduction}
Panoptic Narrative Grounding~\cite{gonzalez2021panoptic} is a new challenging task that locates the target instances of an image corresponding to the text description via binary pixel masks. Its main challenges not only lie in the joint understanding of multi-modal information but also in many-to-many language-vision alignment, \emph{i.e.}, grounding all related instances or amorphous regions mentioned in the text description. This property also makes it different from a similar grounding task called Referring Expression Segmentation (RES)~\cite{hu2016segmentation, yu2018mattnet, ye2019cross, shi2018key}, which segments only one instance per expression. 

\begin{figure}[t]
\centering
\vspace{-0.2cm}
\includegraphics[width=1\columnwidth]{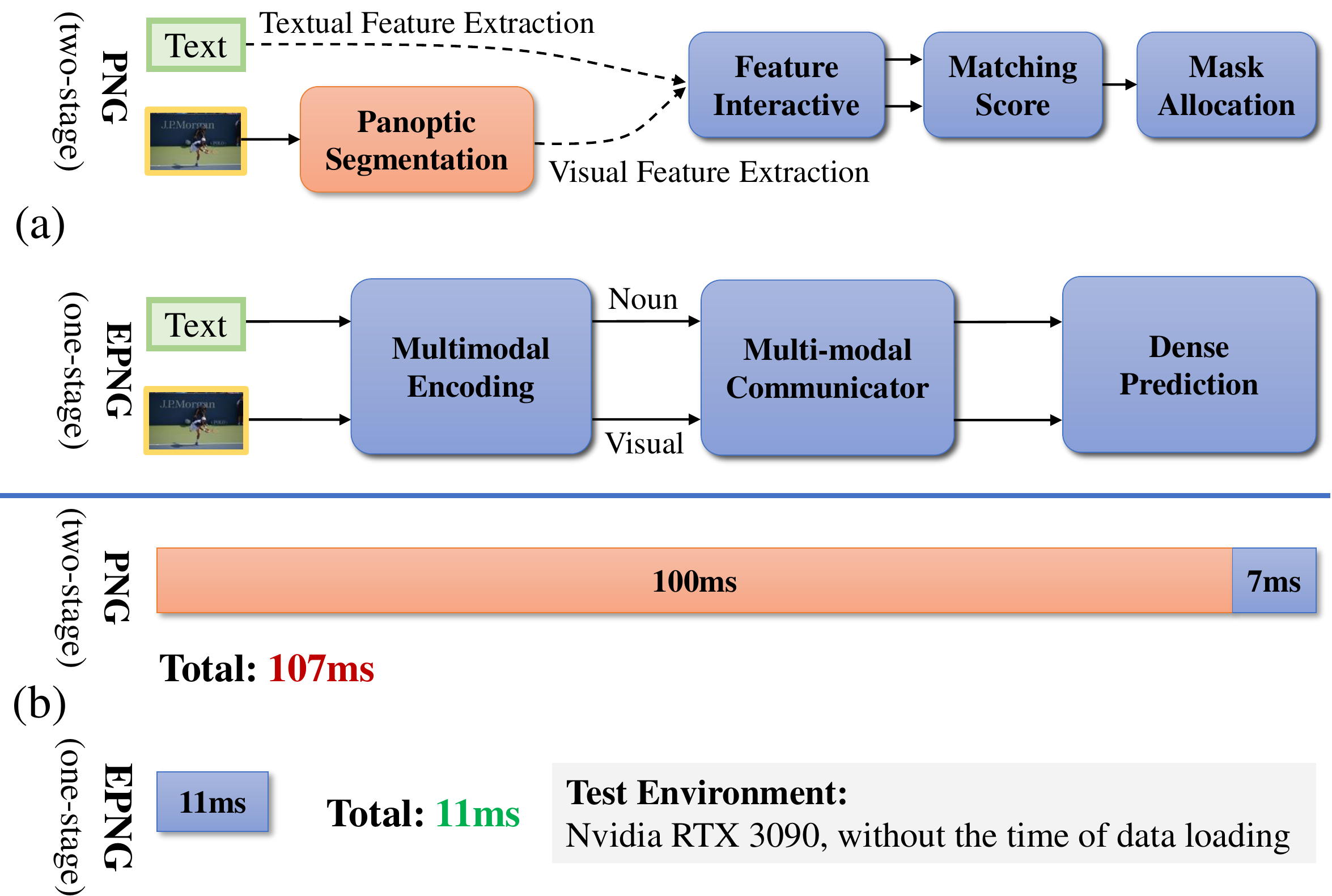} 
 \caption{Comparison of pipeline and inference speed between the proposed EPNG and two-stage PNG. (a) EPNG jointly processes visual and text information to generate referred masks in a one-stage fashion, while PNG relies on mask proposals. (b) Our single-stage EPNG is 10x faster than the two-stage approach, enabling real-time deployment.}
\label{fig1}
\end{figure}

Gonzalez \emph{et al.}~\cite{gonzalez2021panoptic} first explore this task and propose a preliminary solution in a two-stage fashion, as illustrated in Fig. \ref{fig1} (a). First, the pre-trained panoptic segmentation models like PFPN~\cite{kirillov2019panoptic} are used to provide a set of candidate masks of the given image. Secondly, these masks are further transformed into convolution features and then ranked by cross-modal matching. Overall, with the help of panoramic segmentation models, this two-stage solution defines PNG as a mask-text matching problem, greatly reducing the difficulty of prediction. 

However, this solution still suffers from two limitations. On the one hand, such a two-stage approach requires offline feature extraction, storage, and alignment, which is inevitably time-consuming. This limitation poses a huge obstacle to real-time applications, \emph{e.g.}, text-to-image retrieval, and video matting. On the other hand, the pre-trained panoramic segmentation model requires massive mask annotations, which place a greater burden on the already expensive expenditure of PNG. More importantly, the performance of these panoptic segmentation models also limits the upper bound of PNG models.

To solve the above problems, a natural way is to design an efficient single-stage network for end-to-end training from scratch. However, this solution also encounters two challenges that are critical for PNG. First, in PNG, each pixel can be subordinated to different masks, which is greatly different from panoptic segmentation \cite{kirillov2019panoptic}. This property makes the model need to capture visual semantics from macro- to micro-views. However, existing methods only focus on global modeling and overlook local information, resulting in limited performance. Second, PNG involves more complicated relationships than other grounding or segmentation tasks~\cite{liu2017referring, luo2017comprehension, yu2017joint}. In each example, multiple nouns of an expression may correspond to the same mask, or one noun may refer to multiple masks. This case further increases the difficulty of vision-language alignment.

In this paper, we propose a novel End-to-End Panoptic Narrative Grounding network (EPNG) for real-time panoptic narrative grounding, as shown in Fig.~\ref{fig1}. 
Specifically, EPNG adopts a visual encoder to extract the features of the given image, based on which a decoder is deployed to predict masks for different noun phrases.

To enhance local semantic modeling, we introduce Locality-Perceptive Attention (LPA) to enhance grid features via neighborhood interactions based on their spatial priors. In LPA, different attention heads are allowed to perceive visual information in different receptive fields, thus achieving multi-scale modeling. To ensure the semantic consistency of many-to-many relationships in PNG, we design a new bidirectional Semantic Alignment Loss (SAL), which uses one modality as an anchor to eliminate the deviation of similar semantic tokens of the other modality. With these innovative designs, EPNG is superior in cross-modal reasoning while keeping real-time inference.

Conclusively, the contributions of our work are as below:

\begin{itemize}
    \item We propose a real-time End-to-End Panoptic Narrative Grounding network (EPNG), which greatly reduces computation overhead via unifying cross-modal alignment and mask prediction in one forward structure. 
    
    \item We propose two novel designs, namely Locality-Perceptive Attention (LPA) and bidirectional Semantic Alignment Loss (SAL). LPA enhances visual features at different scales to understand complex cross-modal relationships. SAL regularizes the semantic consistency problem by performing contrastive learning between pixels and noun phrases.
    
    \item On the benchmark dataset, EPNG is on par with or even better than existing two-stage methods, while its inference is 10 times faster. In addition, it requires no additional mask annotations for pre-training.  
\end{itemize}

\section{Related Work}

\subsection{Panoptic Segmentation}

Panoptic segmentation aims to entirely understand scenes containing things and stuff. Following the benchmark proposed by~\cite{kirillov2019panopticseg}, the earlier methods treated it as the combination of things masks and stuff masks~\cite{1809.02110}, \emph{e.g.}, PFPN~\cite{kirillov2019panoptic}, Panoptic-DeepLab~\cite{cheng2020panoptic}, and UPSNet~\cite{xiong2019upsnet}. Recently, things and stuff are expected to be treated uniformly~\cite{carion2020end, wang2021max, cheng2021per}. To eliminate the difference between things and stuff, part of those like PFCN~\cite{li2021fully}, K-Net~\cite{zhang2021k}, and Panoptic SegFormer~\cite{li2022panoptic} try to use the kernel to represent things and stuff uniformly and generate masks by the convolution on feature maps, which obtain significant performance. Benefiting from those methods, our model utilizes word features as reliable kernels to get corresponding masks through the convolution on multi-modal features. 
\subsection{Referring Expression Segmentation}

Recently, multi-modal applications have received a lot of attention and made significant progress~\cite{ji2022knowing,ji2021improving,ma2022x,ma2022knowing,zhou2019plenty}. Among them, as a prevalent task in multi-modal communities, Referring Expression Segmentation (RES)~\cite{hu2016segmentation,yu2018mattnet,ye2019cross,shi2018key} is to segment a referent based on the understanding of a related short phrase. In sequential order, previous models~\cite{li2018referring,liu2017recurrent,margffoy2018dynamic} obtain a set of proposals by a general method of segmentation and pick up a better one that is described by the given short phrase. With the strength of leveraging visual information, however, the upper bound of those methods is seriously restricted by the performance of the segmentation models. After that, a batch of methods is developed for refining segmentation masks by a single-stage network~\cite{ye2019cross,liu2019learning,zhou2021real}, which brings higher rates of false positive segmentation. In summary, RES is an incomplete task with the neglect of stuff and many-to-many relationships between natural language and images. Additionally, whether things and stuff or the many-to-many relationships should be considered in Panoptic Narrative Grounding. 

\subsection{Panoptic Narrative Grounding}

The existing method~\cite{gonzalez2021panoptic} handles it with a two-stage paradigm, which first obtains a lot of candidate panoptic masks by a pre-trained panoptic segmentation model. With those candidates, a scoring module is used to assign plural masks to referred phrases. This paradigm achieves impressive performance, nevertheless, the expensive computation cost and space cost on the stage of segment becomes the barrier to real-time. Because of the reasons above, we propose an End-to-End Panoptic Narrative Grounding network (EPNG) to generate the corresponding mask directly from the noun phrases.
\begin{figure*}[t]
\centering
\vspace{-0.2cm}
\includegraphics[width=1.0\textwidth]{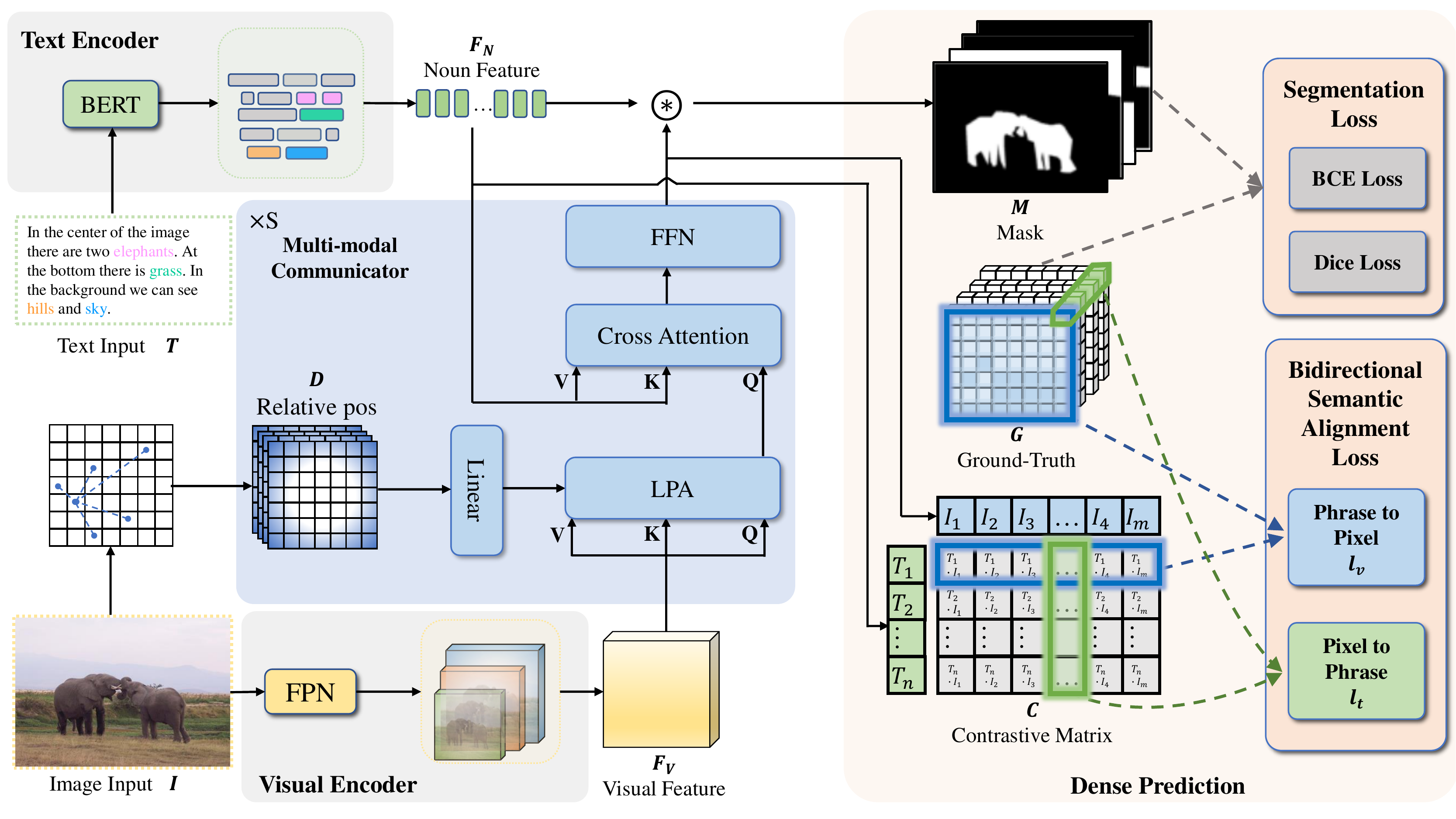}
\caption{The framework of the proposed EPNG. The solid lines denote the pipeline of EPNG, while the dotted lines represent the loss computation during training. During the pipeline, a Multi-modal Encoding module is used to extract the features. Then a Multi-modal Communicator fuses multi-modal features with a Cross Attention module and the proposed LPA. Finally, traditional segmentation loss and the proposed SAL are set to improve the quality of segmentation and align the multi-modal information.}
\label{fig2}
\end{figure*}

\section{End-to-End Panoptic Narrative Grounding Network}

In this section, we give a detailed description of our EPNG, of which the framework is illustrated in Fig.~\ref{fig2}. The input images and descriptions are first processed by the visual and text encoders, respectively. A multi-modal fusion module is further deployed for image-text interaction, based on which a dense prediction head is used to predict masks. 

\subsection{Problem Definition}
Unlike the existing two-stage PNG, the proposed one-stage PNG is free of mask proposals, which generates the mask directly based on the expressions and images. We formulate it as a cross-modal dense prediction task.

Specifically, given an image $\mathbf{I}$ and the corresponding text $\mathbf{T}$, the goal of PNG is to find the nouns $\mathbf{N}=\{n_\ell\}_{\ell=0}^L$ that each pixel $i$ belongs to, where $n_\ell$ is the $\ell$-th noun and $L$ denotes the number of the noun phrases. Then the probability of the obtained mask $\mathbf{M} \in \{0, 1\}$ is formulated as:
\begin{equation}
    p\left(\mathbf{M}\right) = \prod_{i \in \mathbf{I}}\prod_{\ell=0}^L p\left(i|\mathbf{I}, \mathbf{T}, n_\ell\right).
\end{equation}
\subsection{Multi-modal Encoding}

\subsubsection{Visual Encoder}
Given an image $\mathbf{I} \in \mathbb{R}^{H \times W \times 3}$, we first adopt a visual backbone~\cite{lin2017feature} to extract the multi-scale visual features, \emph{e.g.}, $\mathbf{F}_{v1} \in \mathbb{R}^{\frac{H}{8} \times \frac{W}{8} \times C_{1}}$, $\mathbf{F}_{v2} \in \mathbb{R}^{\frac{H}{16} \times \frac{W}{16} \times C_{2}}$, and $\mathbf{F}_{v3} \in \mathbb{R}^{\frac{H}{32} \times \frac{W}{32} \times C_{3}}$ . Then we obtain the final visual feature $ \mathbf{F_{v}} \in \mathbb{R}^ {\frac{H}{16} \times \frac{W}{16} \times C} $ by:
\begin{equation}
    \mathbf{F_{v}} = 
    \text{concat}\left[\text{Down} \left(\mathbf{F}_{v1}\right); \mathbf{F}_{v2}; \text{Up} \left(\mathbf{F}_{v3}\right)\right].
\end{equation}
where $ \text{Up} \left(\cdot\right)$ denotes $ 2 \times$ \emph{upsampling}, $ \text{Down} \left(\cdot\right)$ denotes $ 2 \times$ \emph{downsampling} and $ \text{concat} \left[\cdot\right]$ denotes \emph{feature concatenation}. 

\subsubsection{Text Encoder}
Given a sentence $\mathbf{T}$, we follow \cite{gonzalez2021panoptic} to adopt a pre-trained BERT \cite{kenton2019bert} to extract the word embeddings $\mathbf{F_{T}}=\{v_t\}_{t=0}^{|T|}$, where $v_t$ denotes the embedding of $t$-th word. After that, we filter out the noun phrases according to the annotations given by \cite{gonzalez2021panoptic} and then obtain the phrase features by average-pooling the word embeddings in each phrase. These features are then projected by a linear layer, making their feature dimension consistent with the visual features. As a result, the phrase embedding is denoted as $\mathbf{F_{N}} = \{{f_{n_\ell}}\}_{n_\ell=0}^{L} \in \mathbb{R}^{L \times C}$, where $n_\ell$ represents the $\ell$-th noun phrases, and $L$ is the number of phrases.

\subsection{Multi-modal Communicator}

Based on the visual feature $ \mathbf{F_{v}} $ and the textual feature $\mathbf{F_{N}}$, Multi-Modal Communicator is designed for cross-modal interaction and fusion. It consists of $S$ serial identical layer, and each layer is composed of two modules called Locality-Perceptive Attention (LPA) and Cross Attention (CA).

\subsubsection{Locality-Perceptive Attention}

Similar to self-attention~\cite{vaswani2017attention}, LPA aims to improve the input features via modeling their inter-relationships. As argued in ~\cite{cheng2020cascadepsp}, local information is important for the visual segmentation tasks. Then EPNG, going a step further, presents multi-scale local modeling. Each pixel in an image may belong to different masks at the same time. For example, a pixel in cloth may also belong to a person. However, the standard self-attention treats all tokens in the feature map equally. To this end, we reinforce the role of neighborhood information of each pixel when in attention modeling, following \cite{wu2021transformer,ji2022multi}.

Specifically, in the features $\mathbf{F}^{i}$, the 2D spatial coordinates of the $m$-th and $n$-th vectors are denoted as $\left(x_m,y_m\right)$ and $\left(x_n,y_n\right)$, where the superscript $i$ indicates that the feature map is the output of the layer $i$. Then we calculate the Euclidean Distance between these two coordinates:
\begin{equation}
    \mathbf{D}_{m, n}=\sqrt{(x_m-x_n)^2 + (y_m-y_n)^2},
\end{equation}
where $\mathbf{D} \in \mathbb{R}^{\left(H \times W \right) \times \left(H \times W \right)}$, and we truncate the values in $\mathbf{D}$ with an upper bound 2 to explicitly inject the local receptive information. Afterward, for an attention head $j$ in LPA, we transform distance matrix into a coefficient matrix $\mathbf{R}^j \in \mathbb{R}^{\left(H \times W \right) \times \left(H \times W \right) }$, obtained by:
\begin{equation}
    \mathbf{R}^j=\mathbf{W}^j\mathbf{D}. \\
    \label{eq:r}
\end{equation}
The obtained matrix $\mathbf{R}^j$ is used to re-weight the attention, which is given by:
\begin{equation}
    \mathbf{A}^j = \text{Softmax}\bigg(\frac{(\mathbf{F}^{i}\mathbf{W}^j_Q)(\mathbf{F}^{i}\mathbf{W}^j_K)^T}{\sqrt{d_k}} \otimes \mathbf{R}^j\bigg),
    \label{eq:Attention}
\end{equation}
where the projections $\mathbf{W}^j_Q \in \mathbb{R}^{d \times \frac{d}{h}}$ and $\mathbf{W}^j_K \in \mathbb{R}^{d \times \frac{d}{h}}$ are weight matrices, and $d_k$ is a scaling factor. The subscript $j$ represents the $j$-th head, and the number of heads $h$ is set to 8. $\otimes$ represents an element-wise product. In this way, we naturally embed local information into attention modeling. 

Next, we sum the features using the attention weights to obtain the results for head $j$, and aggregate all the results:
\begin{equation}
    \text{Head}^j = \mathbf{A}^j(\mathbf{F}^{i}\mathbf{W}^j_V),
    \label{eq:vi}
\end{equation}
\begin{small}
\begin{equation}
    \text{LPA}(\mathbf{F}^{i}, \mathbf{F}^{i}, \mathbf{F}^{i}) = \text{concat}(\text{Head}^1, \cdots, \text{Head}^h)\mathbf{W}_O,
    \label{eq:MHA}
\end{equation}
\end{small}
where $\mathbf{W}^j_V\in \mathbb{R}^{d \times \frac{d}{h}}$, and $\mathbf{W}_O\in \mathbb{R}^{d \times d}$ are projection matrices.
\begin{equation}
    \mathbf{F}^{i'}=\text{LN}\left(\text{LPA}\left(\mathbf{F}^{i},\mathbf{F}^{i},\mathbf{F}^{i}\right)+\mathbf{F}^{i}\right),
\end{equation}
where $\text{LN}\left(\cdot\right)$ means the \emph{layer normalization} \cite{1607.06450}, and shortcut connection \cite{he2016deep} is applied after the LPA module.

\noindent \textbf{The difference between LPA and  Multi-Head Attention (MHA)}.
 As shown in Eq.~\ref{eq:Attention}, the definition of LPA is based on Multi-head Attention \cite{vaswani2017attention}, but it still has an obvious difference in principle. MHA treats all tokens in the feature map equally and excels at capturing long-range dependencies. However, local information is inevitably ignored in this process. We introduce the local prior of each grid, which is obtained from the distance matrix to attention modeling. Note that, these local priors can be dynamically adjusted by the coefficient matrix $\mathbf{R}^j$ in Eq.~\ref{eq:r}.

\subsubsection{Cross-modal Attention}
Following~\cite{yu2019deep}, we use Cross-modal Attention for modality interactions:
\begin{equation}
\mathbf{F}^{i+1}=\text{FFN}\left(\text{LN}\left(\text{MHA}\left(\mathbf{F}^{i'},\mathbf{F_{N}},\mathbf{F_{N}}\right)+\mathbf{F}^{i'}\right)\right),
\end{equation}
where $\text{FFN}\left(\cdot\right)$ denotes feed-forward network, and $\mathbf{F_{N}}$ is the noun features.

\subsection{Dense Prediction}
Given the fused feature after multi-modal communicator, $\mathbf{F} \in \mathbb{R}^{\frac{H}{16} \times \frac{W}{16} \times C}$, we upsample it to a tensor shape of $\frac{H}{4} \times \frac{W}{4} \times C$. Afterward, we apply each noun phrase feature as a kernel to convolve $\mathbf{F}$. The final masks $\mathbf{M}$ are obtained by:
\begin{equation}
    \mathbf{M}=\text{Up}\left(\text{Sigmoid}\left(\mathbf{F_N} \ast \mathbf{F}\right)\right),
\end{equation}
where $\ast$ represents convolution operation, and \emph{Sigmoid} transforms the results to $(0, 1)$. After upsampling, we set a threshold to force $\mathbf{M} \in \{0, 1\}$.

\subsubsection{Training loss}
Since panoptic narrative grounding is a segmentation task, we try different \emph{seg.} losses in existing tasks, including BCE loss and Dice loss~\cite{milletari2016v}. For BCE loss, the loss function is formulated as
\begin{equation}
    L_{BCE}=\sum_{\hat{y_i} \in \mathbb{M}}\!-\!\left(y_i \cdot \log \left( \hat{y_i}\right)\!+\!\left(1-y_i\right) \cdot \log \left( 1-\hat{y_i}\right)\right),
    \label{loss:bce}
\end{equation}
where $\hat{y_i}$ is the prediction of the i-th pixel and $y_i$ is the ground-truth. Dice loss is defined by
\begin{equation}
    L_{Dice}=1-\frac{2|M\bigcap G|}{|M|+|G|},
    \label{loss:dice}
\end{equation}
where $M$ is the generated mask and $G$ is the ground truth, the value of which all belongs to $\{0, 1\}$. Considering these loss functions are designed for single-modal tasks~\cite{he2017mask}, we propose a new loss called bidirectional Semantic Alignment Loss (SAL) for regularizing the semantic consistency between modalities.

\subsubsection{Bidirectional Semantic Alignment Loss}
As mentioned above, PNG has complex many-to-many relationships, \emph{i.e.}, a mask may belong to several noun phrases or \emph{vice versa}. However, the above segmentation loss only considers the one-to-one interaction between the phrases and mask, while ignoring the semantic connection between them. Inspired by~\cite{kamath2021mdetr}, we design a SAL to guarantee semantic consistency, which encourages the multi-modal features with the same semantics to be similar.

\begin{table*}[t!]
\centering
\resizebox{1.00\textwidth}{!}{
\begin{tabular}{l|ccccc|ccc|ccc|ccc}
\toprule
\multirow{2}{*}{Method} & \multicolumn{5}{|c|}{Average Recall} & \multicolumn{3}{c}{Inference Time} & \multicolumn{3}{|c}{Params} & \multicolumn{3}{|c}{Training Data}\\

\cline{2-15}
 & All & Thing & Stuff & Single & Plural & Stage-1 & Stage-2 & All & Stage-1 & Stage-2 & All & Stage-1 & Stage-2 & All\\
 \hline
 PNG~\cite{gonzalez2021panoptic} & 55.4 & 56.2 & 54.3 & 56.2 & 48.8 & 100ms & 7ms & 107ms & 21.0M & 240.3M & 261.3M & 1.3M & 0.8M & 2.1M\\

 \hline
 Baseline (ours) & 40.3 & 34.5 &  50.5 & 42.3 & 31.4 & - & - & \textbf{9.5ms} & - & - & \textbf{76.5M} & - & - & \textbf{0.8M}\\
 EPNG (ours) & 49.7 & 45.6 & 55.5 & 50.2 & 45.1 & -& -& 11ms& - & - & \textbf{76.5M} & - & - & \textbf{0.8M}\\
 EPNG$^\ast$ (ours) & \textbf{58.0} & \textbf{54.8} & \textbf{62.4} & \textbf{58.6} & \textbf{52.1} & - & - & 11ms & - & - & \textbf{76.5M} & - & - & 2.1M\\
 \bottomrule
\end{tabular}}
\caption{Comparison of the EPNG and the previous two-stage method. Baseline means the same design as EPNG except for LPA and SAL.
EPNG$^\ast$ is trained with the same data as PNG.}
\label{table1}
\end{table*}
\begin{table}[t]
\centering
\resizebox{1.00\columnwidth}{!}{
\begin{tabular}{l|ccccc}
\toprule
\multirow{2}{*}{Communicator} & \multicolumn{5}{c}{Average Recall} \\

\cline{2-6}
 & All & Thing & Stuff & Single & Plural\\
 \hline
 w/o PE & 42.8 & 36.5 & 52.1 & 44.2 & 32.5\\
 PE~\cite{vaswani2017attention} & 46.7 & 43.0 & 52.9 & 47.6 & 42.4\\
 SPE~\cite{liu2021swin} & 46.1 & 41.8 & 52.2 & 46.6 & 42.2\\
 DPE~\cite{zhu2020deformable} & 45.9 & 41.5 & 52.0 & 46.5 & 40.3\\
 RPE~\cite{dosovitskiy2020image} & 45.3 & 42.1 & 52.2 & 46.9 & 41.5 \\
 LPA & \textbf{49.7} & \textbf{45.6} & \textbf{55.5} & \textbf{50.2} & \textbf{45.1} \\
 \bottomrule
\end{tabular}}
\caption{Ablation study of the LPA module, where
``w/o PE'' denotes no position embedding, and ``SPE'', ``DPE'', and ``RPE'' mean different relative position embedding from other methods.}
\label{table2}
\end{table}
\begin{table}[t]
\centering
\resizebox{1.00\columnwidth}{!}{
\begin{tabular}{l|ccccc}
\toprule
\multirow{2}{*}{Loss} & \multicolumn{5}{c}{Average Recall} \\

\cline{2-6}
 & All & Thing & Stuff & Single & Plural\\
 \hline
 BCE + Dice & 43.7 & 38.5 & 49.6 & 43.7 & 37.6\\
 BCE + Dice + SAL & \textbf{49.7} & \textbf{45.6} & \textbf{55.5} & \textbf{50.2} & \textbf{45.1} \\
 \bottomrule
\end{tabular}}
\caption{Ablation study of the SAL.
}
\label{table3}
\end{table}

\begin{table}[t]
\centering
\resizebox{1.00\columnwidth}{!}{
\begin{tabular}{c|c|c|cccc}
\toprule
 \multicolumn{2}{c|}{Dataset} & Type & mIoU & p@0.3 & p@0.4 & p@0.5\\
 \hline
 \multirow{4}{*}{RefCOCO} & \multirow{2}{*}{testA} & random & 9.2 & 2.5 & 0.8 & 0.1 \\
 \cline{3-3}
  & & zero-shot & 28.0 & 42.5 & 26.5 & 12.4\\
  \cline{2-7}
  & \multirow{2}{*}{testB} & random & 9.8 & 3.3 & 1.3 & 0.3 \\
 \cline{3-3}
  & & zero-shot & 17.7 & 22.4 & 13.4 & 7.8\\
  \hline
 \multirow{4}{*}{RefCOCO+} & \multirow{2}{*}{testA} & random & 9.3 & 2.6 & 0.7 & 0.2 \\
 \cline{3-3}
  & & zero-shot & 27.7 & 41.1 & 25.6 & 11.9\\
  \cline{2-7}
  & \multirow{2}{*}{testB} & random & 10.5 & 3.8 & 1.4 & 0.3 \\
 \cline{3-3}
  & & zero-shot & 20.6 & 27.4 & 16.3 & 9.4\\
\hline
 \multirow{2}{*}{RefCOCOg} & \multirow{2}{*}{test} & random & 9.9 & 4.0 & 1.5 & 0.6 \\
 \cline{3-3}
  & & zero-shot  & 27.4 & 40.1 & 27.3 & 16.3\\
 \bottomrule
\end{tabular}}
\caption{Zero-shot results of EPNG on RES. EPNG is not trained with RES data.  We average the IoU of every case as the mIoU.}
\label{table4}
\end{table}

Specifically, we first adopt noun phrases as anchors to improve the semantic consistency within visual features. For $i$-th noun phrase $\mathbf{F}_N^i \in \mathbb{R}^C$ and the ground-truth $G^i \in \{0, 1\}^{H \times W}$, the collection of pixels with the class of ``1'' is considered as the positive set, while the one of ``0'' is gathered as the negative set. By increasing the similarity within the positive set, multi-modal information is forced to be aligned. Considering all nouns together, the loss is introduced as follows:
\begin{small}
\begin{equation}
    l_v = \!\frac{1}{L} \!\sum_{i=0}^{L} \!\frac{1}{|G^+|} \!\sum_{j \in G^+} \!-\!\log \!\left(\frac{\exp\left({\mathbf{F}_n^i \cdot \mathbf{F}^j / \tau}\right)}{\sum_{k \in G} \exp\left({\mathbf{F}_n^i \cdot \mathbf{F}^k / \tau}\right)}\right),
    \label{loss:clv}
\end{equation}
\end{small}where $G^+$ is the positive set, denotes the class of ``1'' in the ground truth, and $G$ is the ground-truth. $\tau$ is a temperature coefficient.

Next, We adopt pixel features as anchors to improve the semantic consistency within noun features. Similarly, the loss is introduced as follows:
\begin{small}
\begin{equation}
    l_t \!= \!\frac{1}{|G|} \!\sum_{i=0}^{|G|} \!\frac{1}{|T^+|} \!\sum_{j \in T^+} \!-\!\log \!\left(\!\frac{\exp\left({\mathbf{F}^i \!\cdot\! \mathbf{F}_n^j/ \tau}\!\right)}{\sum_{k \in T}\exp\left({\mathbf{F}^i \!\cdot\! \mathbf{F}_n^k / \tau}\right)}\right),
    \label{loss:clt}
\end{equation}
\end{small}where $T^+$ is the positive gather of the noun set, and $T$ is the whole set, where "1" denotes the pixel belonging to this noun phrase. We combine Eq. \ref{loss:clv} and Eq. \ref{loss:clt} as the SAL loss function. 

During the training, we use the summation of Dice loss, BCE loss, and SAL:
\begin{equation}
    L = \lambda_{1}L_{BCE} + \lambda_{2}L_{Dice} + \lambda_{3}L_{SAL},
\end{equation}
where $\lambda_1, \lambda_2$ and $\lambda_3$ are the hyper-parameters. 

\begin{figure*}[t]
\centering
\vspace{-0.2cm}
\includegraphics[width=1.0\textwidth]{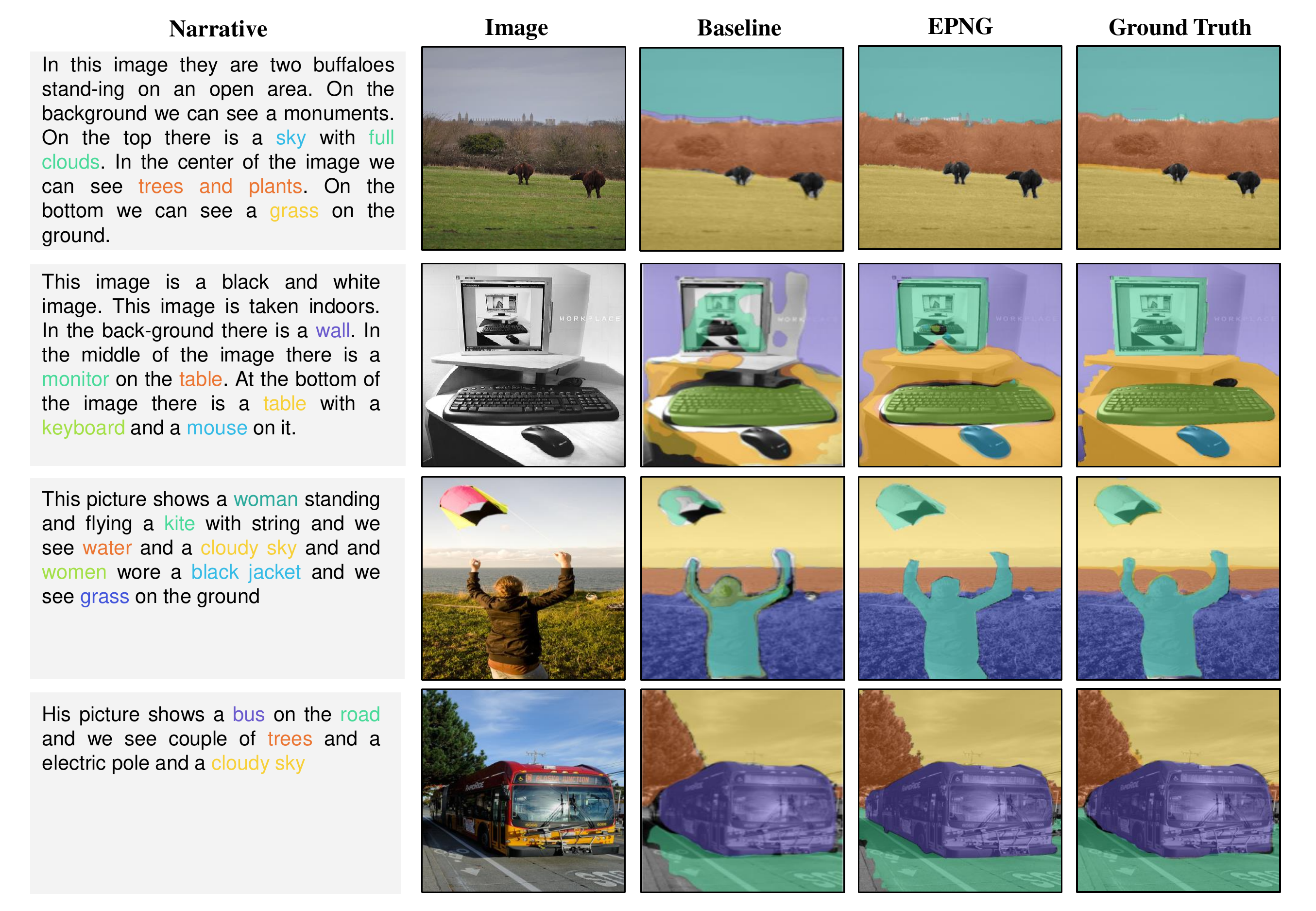}
\caption{Visualization of EPNG. We mark the same color between the nouns in the narrative and the referred pixels.}
\label{fig3}
\end{figure*}

\begin{figure*}[t!]
\centering
\vspace{-0.2cm}
\includegraphics[width=1\textwidth]{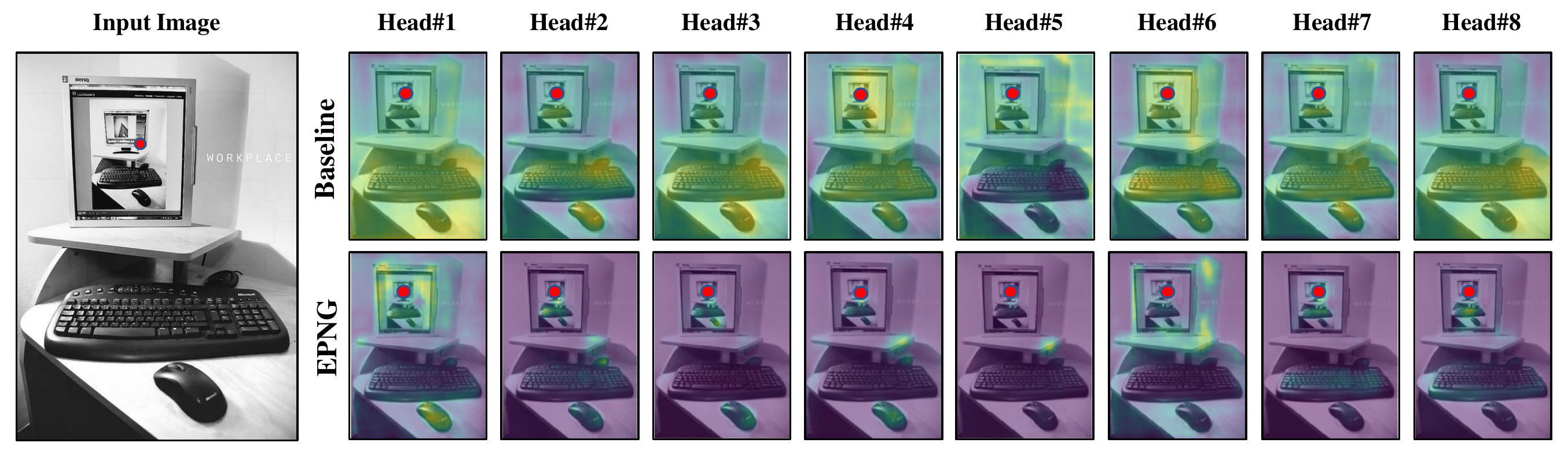}
\caption{Comparison of the attention of the sampled point for the LPA and MHA from the top layer. The red point represents the sampled point.}
\label{fig4}
\end{figure*}

\section{Experiment}
\subsection{Datasets}
We train and compare our model with the existing method on the Panoptic Narrative Grounding dataset \cite{gonzalez2021panoptic}. It is consist of images and the corresponding text. Unlike the brief phrase in other datasets such as RefCOCO \cite{yu2016modeling}, the texts of PNG are long and are a narrative of all items in the complete image and their relationships. It often has hundreds of words and more complex semantic information. The noun-level segmentations are provided for each text. It encompasses both the thing and the stuff, similar to panoptic segmentation.
The difference is that the thing will include both the singular and the plural according to the semantics, which also brings more difficulty for vision-text alignment. The dataset includes a total of 133,103 training images and 8,380 test images with 875,073 and 56,531 segmentation annotations, respectively. 

\subsection{Implementation Details}
\subsubsection{Experimental Settings }
In this paper, we follow \cite{gonzalez2021panoptic} to use the ResNet-101 as our visual backbone, which is pre-trained on the ImageNet \cite{krizhevsky2012imagenet}. BERT is used as the text backbone. During the training process, all the backbones are frozen except for the last two layers of ResNet-101. For parallel training, we increase the input image resolution to $640 \times 640$, so the shapes of the last three layers are $20 \times 20 \times 256$, $40 \times 40 \times 256$, and $80 \times 80 \times 256$, respectively. Moreover, the dimension of text features is $768$. The number of attention heads is 8 and the hidden dimension is 2048. Besides, the number of Layers $S$ is 3. In terms of hyper-parameters, we use $\lambda_{1}=2$, $\lambda_{2}=2$ and $\lambda_{3}=1$ to balance the final loss. We set the initial learning rate $\eta=1e^{-5}$ which is half decayed by every 5 epochs, and fix $\eta = 5e^{-7}$ after 10 epochs. The batch size is 32. We train it on 4 RTX3090 GPUs, which cost 20 hours in total. The optimizer is Adam.

\subsubsection{Metrics}
Following \cite{gonzalez2021panoptic}, we use Average Recall~\cite{gonzalez2021panoptic} as our metric. Specifically, we calculate Intersection over Union (IoU) between masks and the ground truth. We use the integral of the IoU curve as the final metric. Additionally, we simultaneously analyze this measure for the thing, stuff, single, and plural.

\subsection{Quantitative Analysis}
\subsubsection{Comparison with the state-of-the-arts.}
We first evaluate the overall performance of the model using the Average Recall metric, as shown in Tab.~\ref{table1}. In Tab. \ref{table1}, we also introduce a baseline model for comparison, which is the same as EPNG except for LPA and SAL. The performance of the one-stage baseline is much inferior to the two-stage PNG. It is because single-stage models are trained end-to-end from scratch, suffering from greater training difficulties. After adding LPA and SAL, the proposed EPNG can bring up to $23.3\%$ gain on all the masks, $32.2\%$ on the thing, and $9.9\%$ on the stuff, respectively. This fully demonstrates the effectiveness of the proposed method.
In terms of the inference speed, we set the batch size to 12, and calculate the average inference time on each image. Our EPNG has a significant benefit, being 10$\times$ faster than the two-stage model and using only 38\% of its parameters, allowing for model deployment on edge devices.

To make a fair comparison with PNG, we adopt FPN with a ResNet-101 backbone pre-trained with Panoptic Feature Pyramid Network~\cite{kirillov2019panoptic} on MS COCO.
Our EPNG achieves better performance (\emph{i.e.}, 2.6 points of performance gain) at 10$\times$ inference speed, which better demonstrates the contribution of our EPNG.

\subsubsection{Ablation Study}
To verify the contribution of our proposed LPA and SAL, we conduct ablation experiments on the two modules, respectively.
In Tab. \ref{table2}, we compare LPA with other position embeddings. As can be seen that the method that does not use any location and distance information performs the worst, demonstrating the importance of location information for this task. Meanwhile, our LPA achieves the best performance than other relative position embeddings, which are widely used to capture location information \cite{dosovitskiy2020image, liu2021swin, zhu2020deformable}.  


\begin{figure}[t]
\centering
\vspace{-0.2cm}
\includegraphics[width=1\columnwidth]{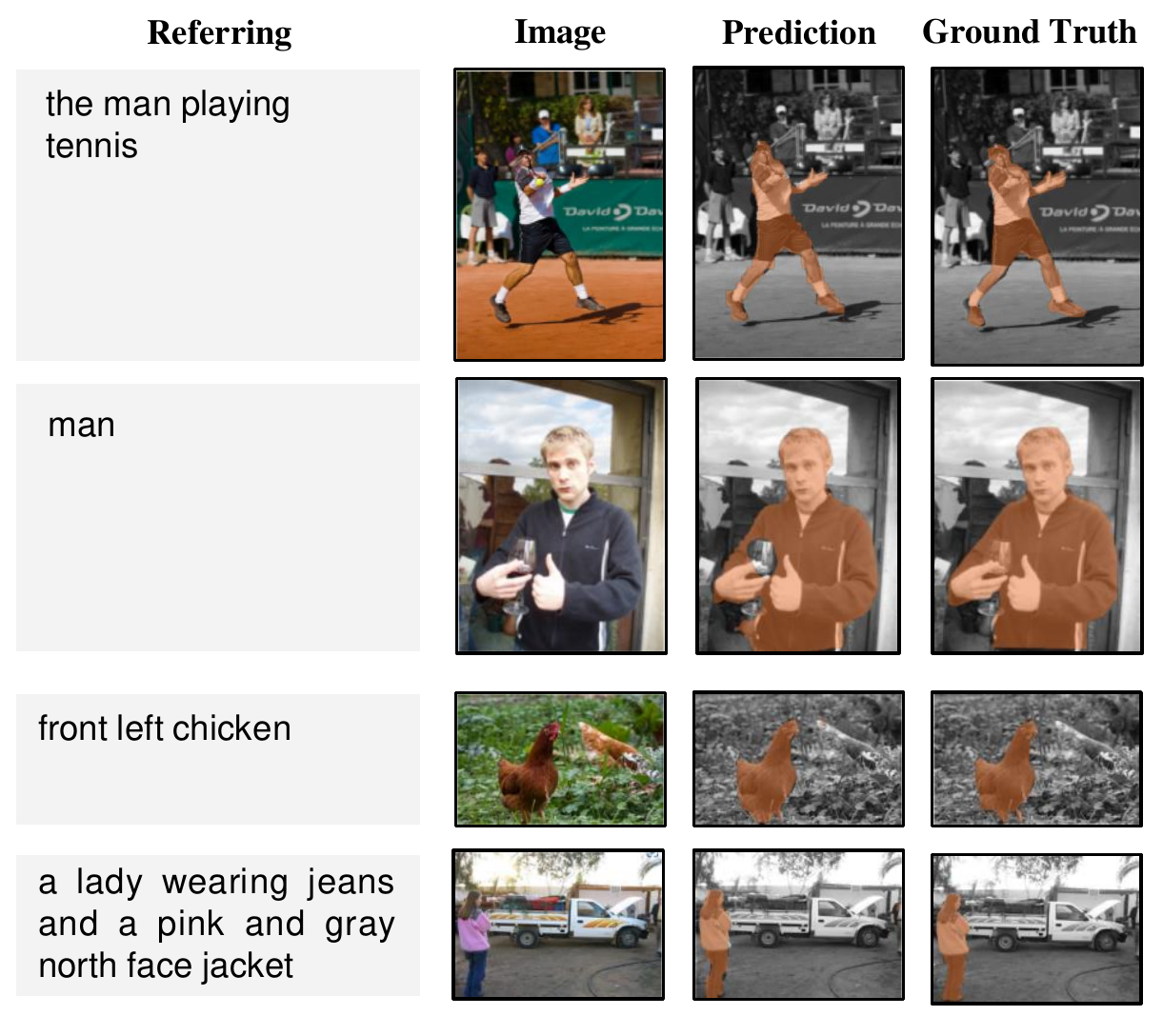} 
 \caption{The visualization of zero-shot setting for RES. }
\label{fig5}
\end{figure}

To verify the efficiency of SAL, we perform two experiments with and without it, of which results are given in Tab.~\ref{table3}. It can be seen that SAL brings a significant improvement of 6.0 points to EPNG, which fully validates our motivation that the improvement of semantic consistency can well improve the segmentation accuracy in PNG.

\subsubsection{Zero-Shot Study for RES}
Meanwhile, we validate our model's generalization by conducting zero-shot experiments on the datasets of the RES task, \emph{e.g.,} RefCOCO \cite{yu2016modeling}, RefCOCO+ \cite{yu2016modeling}, and RefCOCOg \cite{mao2016generation, nagaraja2016modeling}. These datasets are built based on MS COCO~\cite{lin2014microsoft} and each image has a phrase that refers to an object in the image. We use the feature of the whole phrase as the text feature. The results are shown in Tab.~\ref{table4}. By comparing with the fully randomized model, we can find that the zero-shot performance of EPNG is significantly improved. Even on the more complex dataset \emph{i.e.,} RefCOCO+, our performance is close to some early supervised REC models, like \cite{liu2017recurrent}, mIoU of which is 30.48 and 29.5 on the testA and testB of RefCOCO+, respectively.

\subsection{Qualitative Analysis}
\subsubsection{Visualization}

As shown in Fig.~\ref{fig3}, we present some typical grounding results from EPNG compared to the ground truth. 
Compared to the baseline, our method can generate more accurate masks, especially for the edge parts. This further proves the effectiveness of our method. 

\subsubsection{Attention Visualization}

To figure out the role of LPA in our proposed method, we visualize its attention weights during inference in Fig. \ref{fig4}. Compared with the baseline, LPA presents diverse local attention patterns with miscellaneous scopes, which brings a powerful capability to model local semantic relationships. This also validates the motivation of LPA, and our proposed method does allow the model to focus locally and improve the accuracy of attention.

\subsubsection{Zero-Shot for RES}

Additionally, Fig.~\ref{fig5} shows some qualitative results of our zero-shot study on RES. Our method can achieve very accurate segmentation, which fully demonstrates its transferability. For example, in the second case, our model identifies the segmentation of ``man'', which is wrong even in the ground truth. Due to the ability of the finest-grained and complex semantic understanding, EPNG can handle more general scenarios with greater potential.  

\section{Conclusion}

In this paper, we proposed an End-to-End Panoptic Narrative Grounding network (EPNG) for real-time inference. To better handle the many-to-many relationships between pixels and phrases, two innovative designs are proposed, namely Local-Sensitive Attention (LPA) and bidirectional Semantic Alignment Loss (SAL), respectively. Extensive experiments show that our proposed EPNG achieves significant performance gains compared to the baseline. More importantly, compared to the two-stage model, EPNG achieves competitive performance with faster inference (10x) and fewer parameters. Furthermore, we conducted zero-shot experiments on RES and achieved surprising performance. These results demonstrate the excellent generalization of our model and also provide a reference for a subsequent unified vision-language segmentation framework. 

\section{Acknowledgments}
This work was supported by the National Science Fund for Distinguished Young Scholars (No. 62025603), the National Natural Science Foundation of China (No. U21B2037, No. U22B2051, No. 62176222, No. 62176223, No. 62176226, No. 62072386, No. 62072387, No. 62072389, No. 62002305 and No. 62272401), Guangdong Basic and Applied Basic Research Foundation (No. 2019B1515120049), and the Natural Science Foundation of Fujian Province of China (No. 2021J01002,  No. 2022J06001).

\bibliography{aaai23}

\end{document}